\documentclass[conference]{IEEEtran}
\ifCLASSINFOpdf
   \usepackage[pdftex]{graphicx}
\else
\fi
\usepackage{amsmath}
\usepackage{amsfonts}

  \usepackage[caption=false,font=normalsize,labelfont=sf,textfont=sf]{subfig}

\begin{document}
%
\title{Poisson Image Denoising Using Best Linear Prediction: A Post-processing Framework}



%
\author{\IEEEauthorblockN{Milad Niknejad, M\'ario A.T. Figueiredo}
\IEEEauthorblockA{Instituto de Telecomunica\c{c}\~oes, and \\
Instituto Superior T\'ecnico, Universidade de Lisboa, Portugal\\}}


\maketitle

\begin{abstract}
In this paper, we address the problem of denoising images degraded by Poisson noise. We propose a new patch-based approach based on best linear prediction to estimate the underlying clean image. A simplified prediction formula is derived for Poisson observations, which requires the covariance matrix of the underlying clean patch. We use the assumption that similar patches in a neighborhood share the same covariance matrix, and we use off-the-shelf Poisson denoising methods in order to obtain an initial estimate of the covariance matrices. Our method can be seen as a post-processing step for Poisson denoising methods and the results show that it improves upon several Poisson denoising methods by relevant margins.
\end{abstract}

%
\IEEEpeerreviewmaketitle

\section{Introduction}
Image denoising in the presence of Poissonian observations is an important problem, appearing in several application areas such as astronomy and medical imaging, where photon-limited images are very common.

In Poisson denoising, each pixel in the noisy image is a realization of a Poisson random variable with  expected value equal to the true underlying pixel to be estimated.  Some Poisson image denoising methods apply a nonlinear \textit{variance stabilization transform} (VST), such as the Anscombe transform \cite{1948_anscombe_transformation}, to the noisy image, in order to approximately transform the noise into Gaussian-distributed. The resulting image is then processed using a denoising method for Gaussian noise; finally, an inverse of the transform is applied to the output of denoising method, in order to obtain the final image estimate. Some methods have been proposed to improve the accuracy of this inverse transform \cite{2011_makitalo_optimal}. However, VST-based methods are less accurate in regimes of low \textit{signal-to-noise ratios} (SNR), which led to the development of methods that directly tackle the Poissonian image.  Two recent examples of such denoising methods are \textit{Poisson non-local means} (PNLM) \cite{2010_deledalle_poisson} and \textit{non-local PCA} (NL-PCA) \cite{2014_salmon_poisson} (more references to earlier work can be found in \cite{2010_deledalle_poisson} and \cite{2014_salmon_poisson}). Approaches that directly deal with the Poissonian nature of the observations have also been proposed for image deconvolution under Poisson noise (see \cite{Figueiredo2010,Harmany2012}, and references therein).

The underlying assumption in many image restoration methods is that the clean patches lie in a low dimensional subspace. Whereas in the Gaussian noise case, this assumption can be used efficiently \cite{2014_gu_weighted,Niknejad_TIP2015}, in the Possionian case, applying this assumption as a prior leads to an intractable optimization problem due to the non-quadratic log-likelihood. To circumvent this difficulty, some methods, such as NL-PCA and the sparsity-based method in \cite{2016_rond_poisson}, assume the patches are represented as the exponential of some data lying in a low-dimensional subspace. However, the exponential of low-dimensional data is not guaranteed to be low-dimensional.  Recently, approaches based on sampling have been proposed to directly approximate the \textit{minimum mean squared error} (MMSE) estimate \cite{pyatykh_mmse_2015}. Although these approaches are generally computationally complex for generic images, it has been shown that it can be effectively applied in a class-specific setting \cite{NiknejadPoisson2017}.

In this paper, we propose a new method based on the \textit{best linear prediction} (BLP) principle, in order to estimate the clean patches as a linear transformation of the noisy ones. This estimator is independent of the noise distribution and it requires merely the covariance matrix of the underlying clean vector. We assume that similar patches in a neighborhood of a reference patch share the same covariance matrix. In order to estimate this covariance matrix,  we use an off-the-shelf Poisson denoising algorithm. Thus, our method can be seen as a post-processing step for Poisson denoising methods. The experimental results reported in this paper show that our method  improves over state-of-the-art denoising methods by relevant margins.

In the following sections, we first review the BLP principle; then, the proposed method is described. Finally, experimental results are reported and conclusions are drawn.

\section{Best Linear Prediction -- BLP}
In this paper, we use a uppercase normal font to indicate random vectors and uppercase bold font for deterministic matrices.  Consider a random vector $X \in \mathbf{R}^n$ with unknown probability density function is to be estimated from an observed random vector $Y\in \mathbf{R}^m$ as an affine function of the form
\begin{equation}
\widehat{X}=\mathbf{B}Y+\mathbf{a},
\label{eq:blp}
\end{equation}
where $\mathbf{B}\in \mathbf{R}^{n\times m}$ and $\mathbf{a}\in \mathbf{R}^n$ are a fixed matrix and vector, respectively. Let $\boldsymbol{\mu}_{{y}}$ and $\boldsymbol{\mu}_{{x}}$ be the mean of the random variables $Y$ and $X$, respectively. Based on the BLP theorem \cite{searle_variance_2009}, the optimal choice for $\mathbf{B}$ and $\mathbf{a}$ in the MMSE sense is well-known to be
\begin{equation}
\mathbf{B}=\mathbf{\Sigma}_{xy}\mathbf{\Sigma}_{yy}^{-1}, \ \mathbf{a}=\boldsymbol{\mu}_{x}-\mathbf{\Sigma}_{xy}\mathbf{\Sigma}_{yy}^{-1}\, {\boldsymbol{\mu}_{{y}}}
\label{eq:coeblp}
\end{equation}
where $\mathbf{\Sigma}_{xy}$ and $\mathbf{\Sigma}_{yy}$ are the cross-covariance matrix between $X$ and $Y$, and the covariance matrix of ${Y}$, respectively. A remarkable feature of BLP is that it is independent of the distribution of the random variables involved, depending only on the aforementioned matrices and mean vectors. In the case of Gaussian noise, it can easily be shown that BLP is equivalent to using a multivariate Gaussian prior with covariance $\Sigma_{xx}$, which is often used in image denosing \cite{Niknejad_TIP2015,2012_yu_solving}; in that case, \eqref{eq:coeblp} is not only the best \textit{linear} predictor, but the best predictor overall in the MMSE sense \cite{Scharf}.

In this paper, we propose to use BLP \eqref{eq:blp}--\eqref{eq:coeblp} in the case where $Y$ is a Poisson random vector with the underlying mean $X$ ($\mathbb{E}[Y|X] = X$),  that is, for $m=n$ and 
\begin{equation}
\mathbb{P}(Y = \mathbf{y}| X = \mathbf{x}) = \prod_{j=1}^n \frac{e^{-x_j} x_j^{y_j}}{y_j !},\label{eq:poisson}
\end{equation}
where $\mathbf{y} \in \mathbb{N}_0^n$ and $\mathbf{x} \in \mathbb{R}_+^m$.
The main challenge in this approach will be to obtain the aforementioned covariance matrices and mean vectors, as  discussed in the next section.

\section{The proposed method}
\subsection{BLP from Poisson Observations}
The Poisson conditional distribution of $Y$ in \eqref{eq:poisson} has an important implication in the cross-covariance matrix $\Sigma_{xy}$.  Recall that the cross-covariance matrix is given by
\begin{equation}
\begin{split}
\mathbf{\Sigma}_{xy} =\mathbb{E}_{X,Y}[X{Y}^T]- \boldsymbol{\mu}_{{x}} \boldsymbol{\mu}_{y}^T,
\end{split}
\end{equation}
where $\mathbb{E}_{X,Y}$ indicates the expectation with respect to the joint distribution of $X$ and $Y$. Using the so-called \textit{law of iterated expectation}, the above can be written as
\begin{equation}
\begin{split}
\mathbf{\Sigma}_{xy} & =\mathbb{E}_{X}\bigl[\mathbb{E}_{Y|X}[{X} {Y}^T]\bigr]- \boldsymbol{\mu}_{{x}} \boldsymbol{\mu}_y^T \\ & =\mathbb{E}_{X}({X} {{X}}^T)-\boldsymbol{\mu}_{{x}}\boldsymbol{\mu}_{x}^T \\ 
& = \mathbf{\Sigma}_{xx},
\end{split}
\end{equation}
because $\boldsymbol{\mu}_{{y}} = \boldsymbol{\mu}_{{x}}$ and $\mathbb{E}_{Y|X}[{X} {Y}^T] = X\mathbb{E}_{Y|X} [Y^T] = X X^T$. In conclusion (as is also true for any zero-mean additive independent noise), in the case of Poissonian observations (which is not additive noise), we have that $\mathbf{\Sigma}_{xy}=\mathbf{\Sigma}_{xx}$. 

A particular property of the Poisson distribution (namely that its mean and variance are equal) underlies a simple relationship between $\mathbf{\Sigma}_{yy}$ and $\mathbf{\Sigma}_{xx}$. A similar relationship exists in the case of additive independent  zero-mean noise with known covariance\footnote{For example, for additive zero-mean white noise of known variance $\sigma^2$, we have the well-known relationship $\mathbf{\Sigma}_{yy}=\mathbf{\Sigma}_{yy} + \sigma^2 {\bf I}$.}, but for the Poisson  model \eqref{eq:poisson}, no further information is needed. Using iterated expectation again, 
\begin{equation}
\begin{split}
\label{eq:covy}
\mathbf{\Sigma}_{yy} & =\mathbb{E}_{Y}(Y Y^T)-\boldsymbol{\mu}_{{y}}\boldsymbol{\mu}_y^T \\
& = \mathbb{E}_{X}\bigl[\mathbb{E}_{Y|X}[Y {Y}^T]\bigr]-\boldsymbol{\mu}_{{x}}\boldsymbol{\mu}_x^T  .
\end{split}
\end{equation}
Conditioned on $X$, the components of $Y$ are independent, thus the off-diagonal elements of $\mathbb{E}_{Y|X}[Y Y^T]$ are simply 
\[
(i \neq j) \; \Rightarrow \; \bigl(\mathbb{E}_{Y|X}[Y Y^T]\bigr)_{i,j} = \mathbb{E}[Y_i] \; \mathbb{E}[Y_j] = X_i \, X_j.
\]
Concerning the diagonal elements,
\[
\bigl(\mathbb{E}_{Y|X}[Y Y^T]\bigr)_{i,i} = \mathbb{E}_{Y|X}[Y_i^2] = X_i + X_i^2 ,
\]
because the mean and variance of a Poisson random variable are identical. Consequently, 
\[
\mathbb{E}_{Y|X}[Y Y^T] = X X^T + \mbox{diag}(X_1,...,X_n)
\]
and
\begin{eqnarray}
\mathbb{E}_{X}\bigl[\mathbb{E}_{Y|X}[Y Y^T] \bigr] & = &\mathbb{E}_{X} [X X^T] + \mathbb{E}_{X} [\mbox{diag}(X_1,...,X_n)] \nonumber \\
&=&  \mathbf{\Sigma}_{xx} +  \boldsymbol{\mu}_x\boldsymbol{\mu}_x^T +  \mbox{diag}(\boldsymbol{\mu}_x), \label{eq:semi}
\end{eqnarray}
where $\mbox{diag}(\boldsymbol{\mu}_x)$ denotes a diagonal matrix with the components of $\boldsymbol{\mu}_x$. Finally, plugging \eqref{eq:semi} into \eqref{eq:covy}, yields
\begin{equation}
\mathbf{\Sigma}_{yy}  = \mathbf{\Sigma}_{xx}  + \mbox{diag}(\boldsymbol{\mu}_x),
\end{equation}
which is a natural result, since Poisson noise can be seen as additive independent noise with variance equal to the underlying clean value. 

Finally, the BLP \eqref{eq:blp} of $X$ from Poisson observations $Y$ modeled by \eqref{eq:poisson} is given by
\begin{equation}
\widehat{X} = \boldsymbol{\mu}_{{x}} +\mathbf{\Sigma}_{xx}(\mbox{diag}(\boldsymbol{\mu}_{x})+\mathbf{\Sigma}_{xx})^{-1}(Y-\boldsymbol{\mu}_{{y}}).
\label{eq:blppatch}
\end{equation}

\subsection{Application to Patch-Based Poisson Denoising}
As in classical patch-based methods, the image is divided into overlapping $\sqrt{n} \times \sqrt{n}$ patches \cite{2010_deledalle_poisson,2005_buades_nonlocalmenas,2007_dabov_bm3d}. The patches are denoised in collaboration with similar patches and then returned to the original position in the image, with the multiple estimates of each pixel (due to the overlapping nature of the patches) averaged  to reconstruct the final clean image estimate. Denoting the (vectorized) $i$-th noisy patch as $\mathbf{y}_i$, we estimate the corresponding clean patch $\mathbf{x}_i$ by BLP, \textit{i.e.}, using \eqref{eq:blppatch},
\begin{equation}
\label{eq:blpp}
\widehat{\mathbf{x}}_i=\boldsymbol{\mu}_{{x}}+\mathbf{\Sigma}_{xx} (\mbox{diag}(\boldsymbol{\mu}_{x})+\mathbf{\Sigma}_{xx})^{-1}(\mathbf{y}_i-\boldsymbol{\mu}_{{y}}).
\end{equation}
where $\boldsymbol{\mu}_{x}$ and $\mathbf{\Sigma}_{xx}$ are estimated as describe in the next paragraphs.

Some methods for Gaussian denoising use the same underlying covariance matrix for similar patches in the image \cite{Niknejad_TIP2015,2012_yu_solving}. These approaches can all be seen as inspired by the collaborative filtering idea pioneered in the famous BM3D denoising method \cite{2007_dabov_bm3d}. In this paper, we follow  \cite{Niknejad_TIP2015}, by assuming the same covariance matrix $\mathbf{\Sigma}_{xx}$ for patches that are similar to a reference patch $\mathbf{y}_r$ in a neighborhood thereof.

Estimating $\boldsymbol{\mu}_{x}$ and $\mathbf{\Sigma}_{xx}$ from a set of clean patches would simply amount to computing the sample mean and sample covariance.  In the presence of additive zero-mean white Gaussian noise of known variance $\sigma^2$, it would still be possible to obtain estimates $\boldsymbol{\mu}_{x}$ and $\mathbf{\Sigma}_{xx}$  from noisy patches \cite{Teodoro2015}. However, this is not so simple in the case of Poissonian observations, because the variance of each observation depends on the underlying clean value, which is of course unknown. Here, we propose to use an off-the-shelf Poisson denoising methods to obtain an initial estimate of the clean image, from which  $\boldsymbol{\mu}_{x}$ and $\mathbf{\Sigma}_{xx}$ can be estimated; these estimates are then plugged into  \eqref{eq:blpp} to obtain the final patch estimator.

\subsection{Proposed Algorithm}
Let ${\bf y}$ denote the whole Poisson noisy image. The first step is to use an off-the-shelf Poisson denoising method to obtain a so-called \textit{pilot estimate} $\tilde{\bf x}$. In the experiments reported below, we will use  NL-PC \cite{2014_salmon_poisson}, VST+BM3D, and the recent state-of-the-art method in \cite{2016_azzari_variance}. 

The remaining steps of the algorithm follow closely the macro structure of BM3D  \cite{2007_dabov_bm3d,LebrunIPOL}. Denoting a reference patch of the  pilot estimate as $\tilde{\bf x}_r$, this image is searched within a window of size $N\times N$, centered at $\tilde{\bf x}_r$, for the $k$-nearest patches in Euclidean distance; let ${\mathcal X}_r$ denote the set of patches thus obtained. From this set of patches, their sample mean and sample covariance are obtained and denote as $\boldsymbol{\mu}_{{x}}$ and $\mathbf{\Sigma}_{xx}$. 

Let ${\mathcal Y}_r$ denote the set of patches in the noisy image at the same locations as the patches in ${\mathcal X}_r$. Using $\boldsymbol{\mu}_{{x}}$ and $\mathbf{\Sigma}_{xx}$ obtained as explained in the previous paragraph, all the patches in ${\mathcal Y}_r$ are denoised by BLP, according to \eqref{eq:blpp}. These denoised patches are then returned to their locations, and averaged wherever they overlap.

The algorithm may be repeated $L$ times, using the obtained denoised image as the next pilot estimate.

\begin{table*}[!h]
\centering
\caption{Results of the proposed method in comparison with NLPCA}
\label{tb:nlpca}
\begin{tabular}{c|c|c|c|c|c|c|}
\cline{2-7}
 & \multicolumn{2}{c|}{peak=$2$} & \multicolumn{2}{c|}{peak=$5$}& \multicolumn{2}{c|}{peak=$10$} \\ 
 \cline{2-7}
 & \multicolumn{1}{|c|}{NLPCA} & \multicolumn{1}{|c|}{NLPCA+BLP}  &  \multicolumn{1}{|c|}{NLPCA} & \multicolumn{1}{|c|}{NLPCA+BLP}  &  \multicolumn{1}{|c|}{NLPCA}& \multicolumn{1}{|c|}{NLPCA+BLP}\\   \hline  
 \multicolumn{1}{|c|}{Cameraman}& 20.84 & \bf{21.21}&20.69&\bf{21.56}&20.82&\bf{22.02}  \\ \hline
 \multicolumn{1}{|c|}{House}&22.64&\bf{23.40}&23.42&\bf{24.29}&23.97&\bf{25.14} \\ \hline
 \multicolumn{1}{|c|}{Barbara}&21.47&\bf{21.80}&21.83&\bf{22.19}&22.02&\bf{22.52}\\ \hline
 \multicolumn{1}{|c|}{Lena}&23.64&\bf{24.30}&24.70&\bf{25.35}&24.99&\bf{25.81}\\ \hline
  \multicolumn{1}{|c|}{Average}&22.14&\bf{22.68}&22.66&\bf{23.35}&22.95&\bf{23.87}\\ \hline
\end{tabular}
\end{table*}

\begin{table*}[!h]
\centering
\caption{Results of our proposed method  in comparison with  VST+BM3D}
\label{tb:bm3d}
\begin{tabular}{c|c|c|c|c|c|c|}
\cline{2-7}
 & \multicolumn{2}{c|}{peak=$2$} & \multicolumn{2}{c|}{peak=$5$}& \multicolumn{2}{c|}{peak=$10$} \\ 
 \cline{2-7}
 & \multicolumn{1}{|c|}{VST+BM3D} & \multicolumn{1}{|c|}{VST+BM3D+BLP}  &  \multicolumn{1}{|c|}{VST+BM3D} & \multicolumn{1}{|c|}{VST+BM3D+BLP}  &  \multicolumn{1}{|c|}{VST+BM3D}& \multicolumn{1}{|c|}{VST+BM3D+BLP}\\   \hline  
 \multicolumn{1}{|c|}{Cameraman}&22.04&\bf{22.32}&24.40&\bf{24.79}&26.09&\bf{26.51}  \\ \hline
 \multicolumn{1}{|c|}{House}&24.04&\bf{24.81}&26.86&\bf{27.43}&28.63&\bf{29.15} \\ \hline
 \multicolumn{1}{|c|}{Barbara}&22.09&\bf{22.36}&24.81&\bf{25.19}&26.60&\bf{27.02}\\ \hline
 \multicolumn{1}{|c|}{Lena}&24.37&\bf{24.99}&26.76&\bf{27.33}&28.60&\bf{29.08}\\ \hline
   \multicolumn{1}{|c|}{Average}&23.13&\bf{23.62}&25.70&\bf{26.18}&27.48&\bf{27.94}\\ \hline
\end{tabular}
\end{table*}

\begin{table*}[!h]
\centering
\caption{Results of our proposed method  in comparison with the method in \cite{2016_azzari_variance}}
\label{tb:azari}
\begin{tabular}{c|c|c|c|c|c|c|c|c|}
\cline{2-9}
 & \multicolumn{2}{c|}{peak=$1$} & \multicolumn{2}{c|}{peak=$2$}& \multicolumn{2}{c|}{peak=$5$}& \multicolumn{2}{c|}{peak=$10$}  \\ 
 \cline{2-9}
 & \multicolumn{1}{|c|}{\cite{2016_azzari_variance}} & \multicolumn{1}{|c|}{\cite{2016_azzari_variance}+BLP}  &  \multicolumn{1}{|c|}{\cite{2016_azzari_variance}} & \multicolumn{1}{|c|}{\cite{2016_azzari_variance}+BLP}  &  \multicolumn{1}{|c|}{\cite{2016_azzari_variance}}& \multicolumn{1}{|c|}{\cite{2016_azzari_variance}+BLP}& \multicolumn{1}{|c|}{\cite{2016_azzari_variance}+BLP}& \multicolumn{1}{|c|}{\cite{2016_azzari_variance}+BLP}\\   \hline  
 \multicolumn{1}{|c|}{Cameraman}&21.17&\bf{21.29}&22.34&\bf{22.65}&24.53&\bf{25.19}&26.38&\bf{26.88}  \\ \hline
 \multicolumn{1}{|c|}{House}&23.56&\bf{23.74}&24.74&\bf{25.15}&26.78&\bf{27.74}&28.76&\bf{29.80} \\ \hline
 \multicolumn{1}{|c|}{Barbara}&21.37&\bf{21.56}&22.33&\bf{22.51}&24.57&\bf{25.12}&26.52&\bf{27.31}\\ \hline
 \multicolumn{1}{|c|}{Lena}&24.01&\bf{24.32}&25.27&\bf{25.77}&27.07&\bf{28.01}&28.54&\bf{29.59}\\ \hline
   \multicolumn{1}{|c|}{Average}&22.53&\bf{22.73}&23.67&\bf{24.02}&25.73&\bf{26.51}&27.55&\bf{28.39}\\ \hline
\end{tabular}
\end{table*}

It is clear from the description that, rather than a full self-contained denoising method, our algorithm can be seen as a post-processing step for other Poisson denoising methods. As shown below in the experiments, the proposed approach improves over several state-of-the-art Poisson denoising algorithms by a non-negligible margin.


\section{Results}
In this section, we evaluate the performance of our algorithm using different methods to obtain the \textit{pilot estimate}: NL-PCA \cite{2014_salmon_poisson}, VST+BM3D \cite{2011_makitalo_optimal}, and a recent state-of-the-art method in \cite{2016_azzari_variance}. The parameters used in the proposed method are as follows. Reference patches are selected with step-size $4$ along both the row and the columns of the image. The patches have size $8\times 8$ and the search window is $40\times 40$. The number of selected similar patches to each reference patch is set to $k=30$. finally, we use $L=2$ iterations of the algorithm.

Table \ref{tb:nlpca}, \ref{tb:bm3d}, and \ref{tb:azari} show the results of the proposed method for some benchmark images, in comparison to the methods used to obtain the corresponding pilot estimates. It can be seen that in all examples, the BLP improves the pilot methods for every tested images. The improvements often exceed $0.5$ dB. The average improvement of our method is also noticeable. In these tables, mainly the low SNR values with peak values from $1$ to $10$ are considered. Although the peak value of $10$ in images is sometimes not considered as a low SNR, it is included to show the capability of our method to improve the higher SNR's even when VST methods are quite accurate in these ranges.
\begin{figure}[!b]
\centering
\subfloat[]{\includegraphics[width=.23\textwidth]{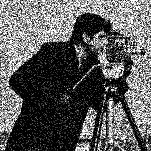}%
\label{(a)}}\\
\subfloat[]{\includegraphics[width=.23\textwidth]{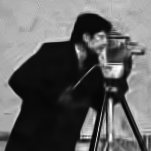}%
\label{(a)}}~
\subfloat[]{\includegraphics[width=.23\textwidth]{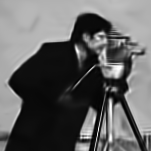}%
\label{b)}}
\\
\hfill
\caption{Example of denoising results of a part of the image Cameraman (a) Noisy image (Peak=$5$), (b) VST+BM3D (PSNR=23.14) (c) Proposed method initialized by VST+BM3D (PSNR=23.51) }
\label{fig:bm3d}
\end{figure}

\begin{figure}[!b]
\centering
\subfloat[]{\includegraphics[width=.23\textwidth]{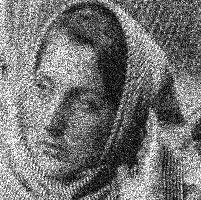}%
\label{(a)}}\\
\subfloat[]{\includegraphics[width=.23\textwidth]{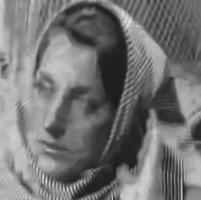}%
\label{(a)}}~
\subfloat[]{\includegraphics[width=.23\textwidth]{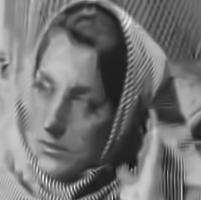}%
\label{(b)}}
\\
\hfill
\caption{Example of denoising of a part of the image Barbara, (a) Noisy image (Peak=$15$), (b) VST+BM3D (PSNR=26.29), (c) VST+BM3D+BLP (PSNR=26.61)}
\label{fig:bm3d2}
\end{figure}

\begin{figure}[!h]
\centering
\subfloat[]{\includegraphics[width=.24\textwidth]{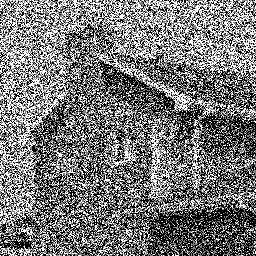}%
\label{(a)}}\\
\subfloat[]{\includegraphics[width=.24\textwidth]{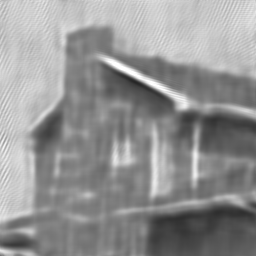}%
\label{(a)}}~
\subfloat[]{\includegraphics[width=.24\textwidth]{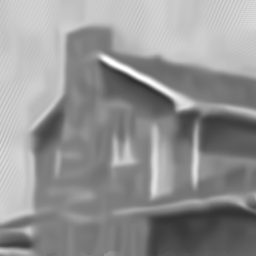}%
\label{b)}}
\\
\hfill
\caption{Example of denoising results of the House image, (a) Noisy image (peak=$2$), (b) NLPCA (PSNR=$24.04$), (c) NLPCA+BLP (PSNR=$24.84$)}
\label{fig:nlpca}
\end{figure}

Some examples of the proposed denoising methods using different initialization are shown and compared to the corresponding initialized images in  Figures \ref{fig:bm3d}, \ref{fig:bm3d2}, and \ref{fig:nlpca}. It can be seen that the proposed technique is able to reduce some of the artifacts produced by the underlying initialization method, yielding visually more pleasing results.

\section{Conclusion}
In this paper, we have proposed a new approach based on \textit{best linear prediction} (BLP) for denoising images contaminated with Poisson noise. BLP is independent of the noise distribution and we showed that, in the Poisson noise case, the only required information is the mean and covariance matrix of the underlying vector being estimated. To implement the BLP approach, our method relies on an off-the-shelf  Poisson denoising method in order to obtain a pilot estimate, from which the required means and covariances are estimated. Consequently, the proposed method can be seen as a post-processing step for Poisson denoising methods. The experiments reported show that our method is able to obtain a noticable improvement over the initial denoised images obtained by several state-of-the-art methods.





%
\bibliographystyle{IEEEbib}
\bibliography{ref}

\end{document}